\documentclass[10pt,twocolumn,letterpaper]{article}

\usepackage{wacv}              
\usepackage[accsupp]{axessibility}
\usepackage{graphicx}
\usepackage{amsmath}
\usepackage{amssymb}
\usepackage{booktabs}

\usepackage[pagebackref,breaklinks,colorlinks]{hyperref}

\usepackage[capitalize]{cleveref}
\crefname{section}{Sec.}{Secs.}
\Crefname{section}{Section}{Sections}
\Crefname{table}{Table}{Tables}
\crefname{table}{Tab.}{Tabs.}

\def\wacvPaperID{2857} 
\def\confName{WACV}
\def\confYear{2025}

\begin{document}

\title{Information Extraction from Heterogeneous Documents without Ground Truth Labels using Synthetic Label Generation and Knowledge Distillation}

\author{Aniket Bhattacharyya, Anurag Tripathi\\
Amazon\\
{\tt\small anikettb@amazon.com,tripaanu@amazon.com}
}
\maketitle

\begin{abstract}
    Invoices and receipts submitted by employees are visually rich documents (VRDs) with textual, visual and layout information. To protect against the risk of fraud and abuse, it is crucial for organizations to efficiently extract desired information from submitted receipts. This helps in the assessment of key factors such as appropriateness of the expense claim, adherence to spending and transaction policies, the validity of the receipt, as well as downstream anomaly detection at various levels. These documents are heterogeneous, with multiple formats and languages, uploaded with different image qualities, and often do not contain ground truth labels for the efficient training of models. In this paper we propose \textbf{T}ask \textbf{A}ware \textbf{I}nstruction-based \textbf{L}abelling (\textbf{TAIL}), a method for synthetic label generation in VRD corpuses without labels, and fine-tune a multimodal Visually Rich Document Understanding Model (VRDU) on TAIL labels using response-based knowledge distillation without using the teacher model's weights or training dataset to conditionally generate annotations in the appropriate format. Using a benchmark external dataset where ground truth labels are available, we demonstrate conditions under which our approach performs at par with Claude 3 Sonnet through empirical studies. We then show that the resulting model performs at par or better on the internal expense documents of a large multinational organization than state-of-the-art LMM (large multimodal model) Claude 3 Sonnet while being 85\% less costly and $\sim$5X faster, and outperforms layout-aware baselines by more than 10\% in Average Normalized Levenshtein Similarity (ANLS) scores due to its ability to reason and extract information from rare formats. Finally, we illustrate the usage of our approach in overpayment prevention.
\end{abstract}

\section{Introduction}
\label{sec:intro}
Visually-rich documents (VRDs) such as invoices and receipts submitted by employees for reimbursement contain a rich mixture of text, images, and layout information. For organizations dealing with such documents, it becomes necessary to properly extract key pieces of information such as vendor name, expense value etc from these documents to protect against abuse, fraud and wastage. Information extraction helps in assessing key factors such as validity of expense, duplicate detection, adherence with spending and transaction policies, and downstream usage such as anomaly/fraud detection through the analysis of aggregated behaviour. These documents are typically \textbf{heterogeneous}, in \textbf{multiple different formats and languages}, and \textbf{may not have ground truth labels} for queries of interest, which means models can not be trained to extract them. 

\begin{figure}
  \centering
    \includegraphics[scale=0.2]{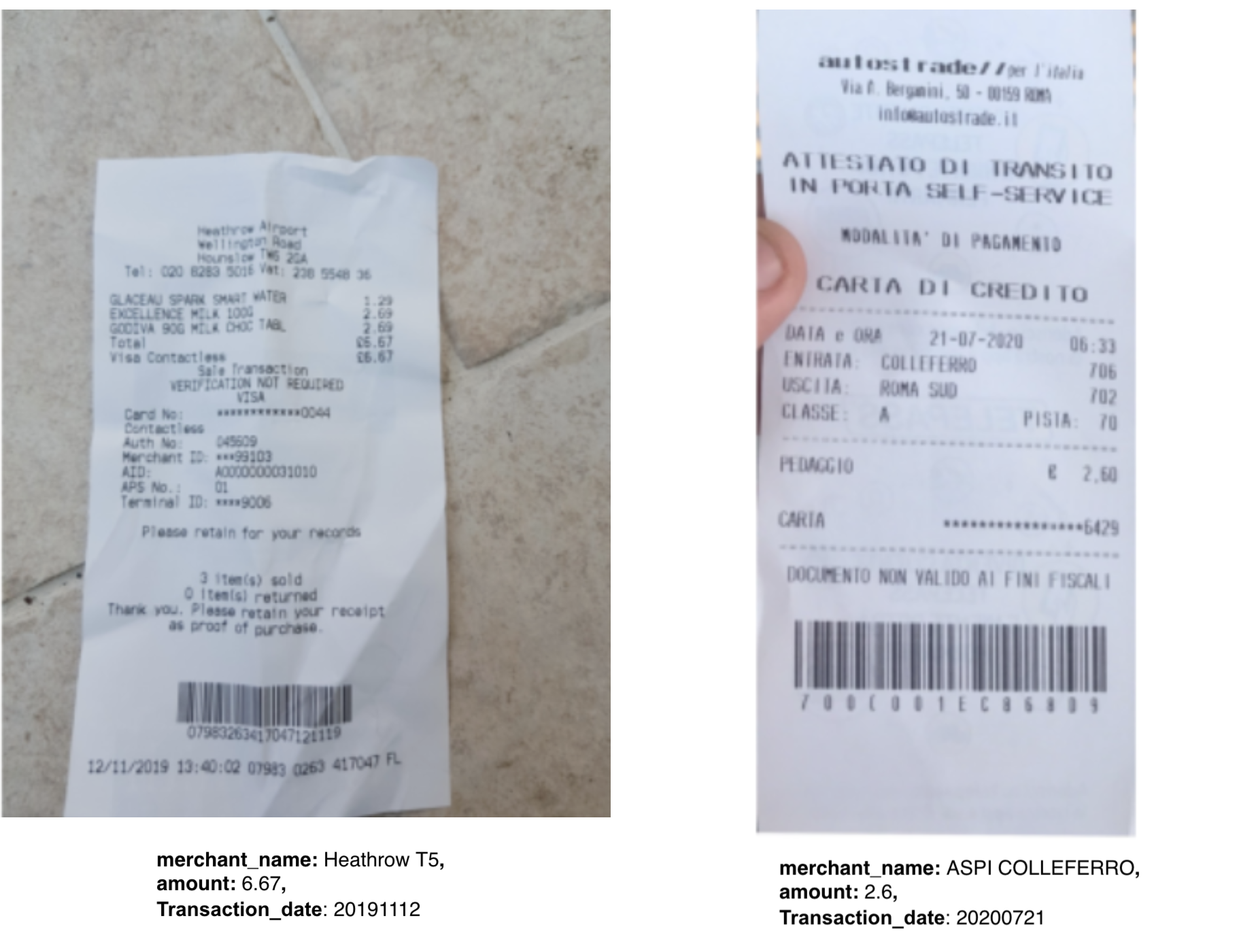}
\caption{Two receipts where the manually typed merchant name does not match the name on the image}
\label{expense}
\end{figure}

Large multinational organizations receive expense reports in the millions annually with supporting invoices/receipts from a large number of vendors in dozens of languages each year. Due to the large number of expense claims, it is not possible to manually verify that the receipts submitted against the expenses are appropriate, i.e. are not duplicates of an older document, or from an unrelated transaction submitted maliciously or erroneously, etc. Employees may be manually required to submit some information, just as justification, merchant name, invoice amount, expense date, etc. However, the value in these fields do not always match the text present in the documents. \cref{expense} shows two receipts submitted by employees of a large multinational organization along with the information present on those documents in expense data. As can be seen, the invoice on the left says Heathrow Airport, while the one on the right says austostrade, whereas the employees have typed in Heathrow T5 and ASPI COLLEFERRO as merchant names respectively.
We refer to these fields (typed in by employees that may not match the documents) as \textbf{noisy labels.}
It may also be of interest to extract information about fields that are not typed in by employees. For such fields, even noisy labels are not available, making it difficult to design solutions to extract them. 

Optical Character Recognition (OCR) is used to extract words from VRDs. Advanced OCR tools such as Amazon Textract and Tesseract are capable of returning words, lines, key-value pairs, tables etc using geometry of recognized text (\cite{tesstract}). However, due to the large number of possible formats of VRDs along with multiple languages, standalone OCRs are not reliable mechanisms to extract information from them consistently. Reasoning is required to understand which part of the document corresponds to the query that is being considered. Contextual clues, such as understanding that a word represents the name of a vendor, or that CGST, VAT, and Umsatzsteuer all represent tax, or that the date mentioned at a certain area of the receipt represents transaction date even if its not labelled as the transaction date, can only be used efficiently by systems that possess the ability to reason.

NLP approaches capable of processing document text and structures have been used to solve these problems. Deep learning approaches such as RNN \cite{aggarwal-etal-2020-form2seq,palm}, CNN \cite{HaoCNN, bertgrid,chargrid} and transformers \cite{wang,majumder} have been leveraged in recent years. These techniques focus on mining information such as structural components of documents. They require extensive labelling of these components, which may not be available in expense receipts, and additionally do not consider the layout of documents. Layout-aware methods such as \cite{layoutlm,layoutlmv3,peng2022ernielayout} use layout information through mechanisms such as bounding box-embeddings to extract information from documents.  However, they need to be pre-trained in tasks involving documents with unseen formats, which is challenging in expense processing cases where labels may not exist.

The recent advancements in the field of large multimodal general-purpose models have provided a new way to approach the problem. Commercial LMMs such as Claude \cite{Anthropic} and ChatGPT \cite{OpenAI} possess the ability to reason and process the query, the document, and provide responses in the desired format. However, these are closed source and are only available through expensive API calls, and even when instruction-tuned, using them in production environments is risky and expensive. Open-source LMM architectures exist which can be fine-tuned and hosted in internal environments. Models such as LLaVa \cite{llava} and CogVLM \cite{cogvlm} etc have shown promising performance in question answering tasks such as SQA, VQAv2. etc. However, these are tasks that involve general descriptions of images or text instead of extraction of particular information and outputting in a precise format. Their zero-shot capabilities are poor, and accurate ground labels are often not available for invoices and receipts for fine tuning.

In this paper, we consider the problem of visually rich document understanding in the absence of labels, and propose a novel solution to leverage the advancements made in large general purpose multimodal models without expensive API calls. We generate synthetic soft labels using Claude 3 Sonnet,  a state-of-the-art large multimodal model (LMM), and train a smaller multimodal model on the synthetic labels to learn from the larger model in a teacher-student architecture. We show that the multimodal model thus trained performs at par with or slightly better than Sonnet under certain conditions, and outperforms other recent solutions such as layout-aware methods. We then show how this model can be used to detect potential abuse, wastage or fraud, by focusing on one particular use case. To summarize, this paper makes the following \textbf{contributions}:
\begin{itemize}
\item We introduce \textbf{T}ask \textbf{A}ware \textbf{I}nstruction-based \textbf{L}abelling (\textbf{TAIL}), a mechanism to generate synthetic ground truth labels for training models on VRDs without labels. We train a multimodal model \textbf{LlaVA-Net}, a VRDU model capable of extracting information in the desired format from expense documents through response-based knowledge distillation without using the teacher model's weights or training dataset.
\item We perform empirical analysis on a benchmark dataset \cite{park2019cord} where ground truth labels are available to demonstrate conditions under which our approach performs near-par with Claude 3 Sonnet on VRD corpuses.
\item  We show that LlaVA-Net achieves performances comparable with instruction-tuned state-of-the-art models like Claude 3 Sonnet while reducing cost by 85\% and being $\sim$ 5X faster, and outperforms layout-aware methods by up to 10\% in ANLS scores in the internal expense dataset of a large multinational organization where only noisy labels are present. 
\item We demonstrate the usage of our approach to detect potential abuse/fraud/wastage through the overpayment detection using mismatch between noisy labels and model output.
\end{itemize}

While in this paper we explore the problems of VRDU in the context of expense documents, the learnings from this paper can be leveraged broadly in structured document understanding problems such as contract and legal document analysis, medical document processing for purposes such as insurance, document processing in banks for automated KYC/underwriting, caption generation, digitization, etc.

\section{Related Work}
\label{relatedwork}
\subsection{VRD understanding}
Previous work in VRD understanding can be categorized into three groups - traditional techniques that use text and vision information, layout-aware models that use visual and layout information through learnable layers on top of NLP, and instruction-tuned and/or fine-tuned large multimodal models capable of reasoning and formatting output. \textbf{Early traditional methods} use rule-based models, or models that learn from handcrafted features \cite{Gorman,jakha,Simon,Marinai,mausam,chiticariu}. These techniques assume form templates are known a priori, or rely on loose rules, and do not generalize to new document templates. \textbf{Deep learning} is leveraged to get more efficient information extraction algorithms. \cite{aggarwal-etal-2020-form2seq} and \cite{palm} used RNNs to extract structural information about documents, such as tables and text fields. Other deep learning frameworks such as CNN \cite{HaoCNN,bertgrid,chargrid} and transformers \cite{wang,majumder} have also been explored. These techniques focus on mining information such as structural components of documents. They require extensive labelling of these components, which may not be available for expense receipts.

\textbf{Layout-aware NLP approaches} have been used in recent years to solve these problems.Transformers have been used by \cite{ DocFormer,hwang2021spatial,bai2022wukongreader,dhouib2023docparser}. Other approaches use language models \cite{devlin2019bert,liu2019roberta,bao2020unilmv2} with additional learnable modules on top that capture layout of documents. \cite{layoutlm} uses 2D position embeddings to incorporate bounding box information of detected words. \cite{layoutlmv2,layoutlmv3,peng2022ernielayout} enhance the usage of position embeddings through a layout-aware attention mechanism. \cite{formnetv2} uses a graph contrastive objective to maximize the agreement of text and image representations. These methods pre-train models on extensive document data such as IIT-CDIP \cite{iitcdip}, and requires further finetuning for document question answering tasks such as DocVQA \cite{docvqa} and FUNSD \cite{funsd}, which have labelled responses. This is a challenge as expense documents have noisy labels at best. 

\textbf{Large multimodal models (LMMs)} have been applied to VRDU as well, recently.  Commercial LMMs such as Claude \cite{Anthropic} and ChatGPT \cite{OpenAI} possess the ability to reason and process the query, the document, and provide responses in the desired format zero-shot, without further fine-tuning. The Claude 3 family of models has been shown in \cite{Anthropicres} to achieve state-of-the-art performances in DocVQA. Open-source LMM architectures have also been developed and tested on question answering tasks. Models such as LLaVa (\cite{llava}) and CogVLM \cite{cogvlm} etc have shown promising performance in question answering tasks such as SQA, VQAv2. etc., when fine-tuned. However, their zero-shot capabilities are poor, and accurate ground labels are not available for invoices and receipts for fine tuning. On the other hand, while commercial LMMs show promising zero-shot performance, they require expensive API calls and have data privacy concerns. 

\subsection{Knowledge distillation}
Response-based knowledge distillation was first proposed by \cite{hinton2015distilling} to compress the knowledge present in ensemble of large neural networks to a single model using the output of the teacher model as soft labels. There have been several variations on knowledge distillation proposed since, typically requiring the original data that was used to train the teacher model \cite{Wangdistil,rastegari2016xnornet,courbariaux2016binarized}. Data-free methods have been proposed since \cite{hu2020creating,ye2020datafree}. However, any distillation method requiring information about the parameters or weights or training dataset of the teacher model can not be applied on closed-source LMMs. Additionally, while knowledge distillation has been studied for models processing images, expense documents are a special type of images where reasoning and layout-awareness plays a key role along with vision.

In this paper, we present \textbf{TAIL}, a method to synthetically generate labels for visually rich documents using a large state-of-the-art multimodal model \textbf{without requiring the training dataset or weights of the teacher model}. We then train a smaller student multimodal model, that we call \textbf{LLaVA-Net} on the labels generated, to teach it to extract fields of interest from expense documents through response-based knowledge distillation. We show that it is able to show at-par performance compared to the teacher model while being less costly and faster, and outperforms recent layout-aware models.


\section{Problem Statement}
Expense receipts and invoices are visually rich documents. Visually rich document understanding (VRDU) is a general task, and in \cref{relatedwork} we discussed VRDU methods and datasets. In this section, we set up the VRDU task using expense document-specific vocabulary, and this can be easily extended to all structured document processing problems.

Given an expense document $E_i, i \in (1,2,..N) $, one may be interested in extracting different attributes $E_{ij}^k$ (vendor name, amount, product purchased, location, etc) with values $E_{ij}^v$, in specific formats. Let $M$ denote the total number of attributes of interest, and $V_j$ denote all possible values for the attribute $E_{ij}^k$. Let $\mathbb V = (V_1 \times V_2....\times V_M)$.

Let the entire corpus of expense documents be characterised by the set $\mathbb E $. Then, the goal of VRDU is to define a mapping $f:\mathbb E\to\mathbb V $ such that 

\begin{equation}
  f(E_i) = (E_{i1}^v,E_{i2}^v,...E_{iM}^v), \forall i \in (1,2,...N).
  \label{eq:important}
\end{equation}

In this paper, we examine two different cases of this problem.

\textbf{Problem 1: Internal ExpenseQA dataset.} This consists of internal receipts and expense reports from a multinational organization, with noisy labels as discussed earlier. To create a benchmark dataset on which different approaches could be assessed, we chose fields for which noisy labels were present. While noisy labels aren't exact, they are the closest fields available for ground truth. In particular, the problem statement formulated was as follows: given an expense document, extract the three following pieces of information - \textbf{merchant name}, \textbf{invoice/receipt amount}, and \textbf{transaction date}. Here, M=1. Additional details are provided in \cref{sec:expenseqa}.

\textbf{Problem 2: CORD dataset.} The CORD dataset \cite{park2019cord} consists of 1000 Indonesian receipts with differing image quality and formats. Crowdsourced exact labels are available for each CORD receipt for 20 $+$ attributes. We use CORD dataset to understand 1) how LLaVA-Net performs on benchmark datasets, and conditions of optimal performance, 2) how training on TAIL labels affects performance with compared to training on actual labels, and 3) impact of image quality on annotation. We report results for M=1 (total amount) and M=24 setup (we leave some sparsely labelled attributes).
\section{Proposed Solution}

\subsection{Task Aware Instruction-based Labelling - TAIL}
\label{label_soln}
Historically, the absence of labels has been tackled by crowdsourcing labels. This is a time-consuming and expensive task. Additionally, expense documents are sensitive containing confidential information such as employee name, which can't be shared with other parties. Instruction Tuned large multimodal models have recently shown promising performance in zero-shot learning on various visual question answering and document understanding tasks.

In \cite{Anthropicres}, the Claude family of models, particularly Claude Sonnet, shows the most consistent performance across the three tasks DocVQA \cite{docvqa}, A12D \cite{a12d} and ChartQA \cite{masry2022chartqa}. Additionally, it is available on Amazon Bedrock, where Anthropic has set up an escrow account without outbound permission \cite{bedrock}. This takes care of the problem of data privacy as well. We chose Sonnet for synthetic label generation.

As discussed earlier, the internal expense receipts do not contain comprehensive labels. To control the output and to provide us with some benchmark for the assessment of our work, we chose formats present in the noisy labels. The three tasks chosen were - extraction of \textbf{merchant name}, extraction of \textbf{invoice/receipt amount}, extraction of \textbf{transaction date}. For CORD dataset, crowdsourced labels are already available for 20$+$ attributes in each receipt. We tested our approach on one attribute (M=1, $E_{i1}^k$ = total amount) as well as 24 attributes, with the same format as in the crowdsourced labels.

To optimize zero-shot learning, prompts need to be efficient. General prompting of the type "What is the invoice amount?" led to responses such as "The invoice amount shown in the image is \$10.46", which does not match the format of the noisy labels or of the images.  Anthropic guidelines \cite{prompt} suggest the usage of examples, controlled output formats, and roles for Claude. Works such as \cite{latin} show that task-specific prompting with roles and examples lead to significant improvements in models such as Claude and GPT. Labels  generated were investigated for \textbf{adherence to requirements}, matches with \textbf{noisy/exact labels}, and with the actual \textbf{expense documents}. Mismatches with noisy labels and images were investigated to understand how to get better responses. The final prompts chosen were \textbf{task-aware instructions}, i.e. contained specific instructions on what to return, and were customized for each specific task. The finalized prompt template, along with the function of each line for M=1, is shown in \cref{tab:TAIL_prompts}). The formats chosen for the internal dataset - were as few words as possible for merchant name, invoice local currency value without currency sign (such as 45 for \$45) for amount, and yyyymmdd for date. For CORD dataset with M=1, the format chosen was the total price as mentioned exactly in the receipt. 
\begin{table*}[t]
	\small
		\label{e}
		\centering
		\begin{tabular}{ll} 
			\toprule
            \textbf{Function} & {Prompt Template}\\
			\midrule
      Role/Task &  You are a seasoned invoice transcriber. Given an invoice image, you return \\
Role/Task &  \color{red}\textit{\{value\_placeholder\}}\\ 
  Example1 &     For example, if the {\color{red}\textit{\{value\_placeholder\}}} is {\color{red}\textit{\{example1\}}}, you return {\color{red}\textit{\{example1\_formatted\}}}.\\ 
 Example2 &      If the image is a starbucks invoice worth \$45 
       and transaction was on 5th April 2023,
            you return\\
  Example2 &          {\color{red}\textit{\{business\_example\_value\}}}.\\
    MissingExample &        If you are not confident about {\color{red}\textit{\{value\_placeholder\}}},\\
  MissingExample &            return None.
            For example if in the same 
            invoice you cant read the {\color{red}\textit{\{value\_placeholder\}}}, \\
      MissingExample &        return None.\\
   Rules &           Only the {\color{red}\textit{\{value\_placeholder\}}}, no extra words.\\
     Request &        Return {\color{red}\textit{\{value\_placeholder\}}} for this invoice  \\
      image data &      \textit{image}\\
		\end{tabular}  
		\caption{TAIL prompt templates, M=1}
    \label{tab:TAIL_prompts}
	\end{table*}

For CORD dataset with M=24, the format chosen for output was JSON, and examples were given for particular fields instead of the entire JSON. This was done to keep prompts succinct, as long prompts have been shown to lead to LLMs ignoring instructions towards the middle of prompts \cite{liu2023lostmiddlelanguagemodels}. We changed the names of the receipt attributes to self-explanatory keys (such as name instead of nm, line\_item instead of menu, etc), as it has been shown that this leads to better performance \cite{perot2024lmdxlanguagemodelbaseddocument}.
\newline \newline Actual prompts for CORD dataset are shown in the supplementary material. We do not share the exact prompts for the internal dataset to maintain confidentiality. 
\newline \newline The labels thus generated were found to adhere to required format, and had an ANLS \cite{ANLS} of 50\%, 70\% and 76\% respectively for the merchant name, amount and date queries with the noisy labels in the internal dataset. On CORD dataset the labels generated with TAIL prompts under M=1 setup showed an ANLS of 94\% with crowdsourced labels, and a tree-edit distance and TED of 93\% under the M=24 setup. These metrics are explained in \cref{Experimental Setup and Results}.

Note that additional good practices exist to improve performance in LMMs. In-context learning (ICL) \cite{brown2020languagemodelsfewshotlearners} has been shown to lead to more relevant generation, especially when some similarity metric is used to include the most similar examples in the prompt. Chain-of-thought (COT) \cite{wei2023chainofthought} can be used in final output as well, to get better conditional annotations in the presence of multiple likely candidates. OCR output can be leveraged inside prompts to augment an LMMs vision capabilities. However, all of these approaches lead to increased latency as well as cost, by increasing the number of tokens inputted and outputted, and by including an OCR layer. In this paper, we show how open-source LMMs can be trained to perform at-par with large state-of-the-art LMMs while decreasing latency (~5X faster) and cost (by upto ~85\%), as cost and latency were both important given the huge volume of expense receipts present without labels. Additional good practices such as ICL, COT, advanced OCR output etc can be then leveraged by practitioners to further improve accuracy of annotations.

\subsection{Student model for Knowledge Distillation: LLaVA-Net}
To effectively learn from TAIL labels, student models need to possess vision, reasoning, and question-answering skills. For reasoning and question answering, open source foundational language model Vicuna \cite{vicunaz} uses pre-trained Llama weights \cite{llama} and fine-tunes on GPT4 conversation to achieve better instruction following capabilities. 

LLaVA (\cite{llava} combines Vicuna with CLIP , a vision encoder capable of state-of-the-art zero-shot textual image representation by training models on (image, caption) samples\cite{CLIP}. The two modules are combined through a learnable projection matrix, and further trained on instruction following datasets. LLaVA also demonstrated multilingual capabilities, which are needed on the ExpenseQA task. Later versions of LLaVa also provided integration with Mistral instead of Vicuna, which has shown superior performance than Llama 13B in various benchmarks \cite{jiang2023mistral7b}. Since Vicuna is based on Llama, we chose the pre-trained weights with Mistral instead of Vicuna. We took the pre-trained LLaVA weights and fine-tuned the text component on TAIL labels using LoRA \cite{lora}. LoRA allows us to finetune the 7B parameter text component using 8 GPUs on SageMaker in 10 hours. The final model trained on TAIL labels using Sonnet is called \textbf{LLaVA-Net}.
\section { Dataset and Experimental Setup}
\label{sec:expenseqa}
\subsection{ExpenseQA Dataset Creation}
TAIL prompts for all three tasks were used on a sample of 11000 expense documents. The final dataset created contained both noisy and TAIL labels for all the tasks along with the tail prompts. This was divided into a training sample of 10000 rows and a validation sample of 1000. This dataset was used to further train LLaVA-Net, and to evaluate LLaVA-Net against benchmarks. \cref{tab:expenseqa} shows the TAIL labels along with the noisy labels for the two receipts shown in \cref{expense}. Notably, the TAIL labels match the text present in the images. While this example focuses on merchant name mismatches, other receipts in the dataset contained mismatches for the two other fields as well.
\begin{table*}[t]
	\small
		\label{e}
		\centering
		\begin{tabular}{lllllll}
			\toprule
            \textbf{Receipt} & \textbf{merchant name} & \textbf{merchant name}  
            & \textbf{amount} & \textbf{amount}
            & \textbf{transaction date}& \textbf{transaction date}\\
                        \textbf{} & \textbf{noisy } & \textbf{TAIL} & \textbf{noisy } & \textbf{TAIL} & \textbf{noisy } & \textbf{TAIL} \\
			\midrule
         1  & Airport T5 &      Heathrow Airport    & 6.67  & 6.67 & 20191112 & 20191112  \\
   
        2& meal Nampa        & Holy Cow!  & 2.6 & 2.60 & 20200721 &20200721  \\

		\end{tabular}
  
		\caption{Noisy and TAIL labels for invoices in \cref{expense}}
    \label{tab:expenseqa}
	\end{table*}
\subsection{CORD Dataset}
CORD \cite{park2019cord} contains 1000 Indonesian receipts, divided into train, validation and test samples of size 800,100 and 100. Along with the images, CORD also contains crowdsourced labels, and OCR output with bounding boxes. However, to simulate expense data without labels, we did not use the OCR output, relying instead on the vision capabilities of models under consideration, and TAIL labels for training. We used the actual labels only for testing, to understand the efficacy of our approach.

\subsection{Experimental Setup and Results}\label{Experimental Setup and Results}
We investigated five research questions to assess the performance and appropriateness of TAIL and LLaVA-Net in the ExpenseQA and CORD datasets. 

In \textbf{research question 1}, We look at the performance of LLaVA-Net on both ExpenseQA and CORD, and compare it with Sonnet as well as layout-aware benchmarks.

In \textbf{research question 2}, we focus on the performance discrepancies between layout-aware methods and LLaVA-Net, and analyse why LLaVA-Net is more effective.

In \textbf{research question 3}, we look at the impact of image blur on the performance discrepancies between layout-aware methods, LLaVA-Net and Sonnet, and demonstrate conditions under which LLaVA-Net performs near-par with Sonnet.

In \textbf{research question 4}, we investigate whether using synthetic TAIL labels biases LLaVA-Net, and makes it prone to repeating the errors of Claude Sonnet. 

In \textbf{research question 5}, we illustrate how LLaVA-Net output can be used to prevent potential fraud/abuse/wastage through the detection of overpayment risk.
\subsubsection{RQ 1: Performance comparison.}

Here, we look at the performance of LLaVA-Net, trained on TAIL labels generated using Sonnet on both ExpenseQA and CORD. We also compare it with the performance of LLaVA - zero shot with TAIL prompt, as well as LLaVA finetuned on noisy labels (exact labels for CORD) instead of TAIL labels, to understand the lift provided by TAIL labels.  We also compare its performance with the performance of layout-aware methods which are state-of-the-art in information extraction. As discussed in the related works section, layout-aware methods need rigorous labelling for fine-tuning, which are not available in ExpenseQA. However, we have generated synthetic TAIL labels, and we used those to fine-tune LayoutLMV3  which has achieved SOTA in question-answering and document processing tasks amongst peers\cite{layoutlmv3}. We did not use LayoutLMV3 on transaction date because TAIL labels for date were designed to match noisy labels instead of the date formats in invoices. We also look at the TAIL labels themselves, and compare those with the noisy labels. 

For CORD, we look at performance under both M=1 and M=24 setups. We compare LayoutLMV3 with Sonnet and LLaVA-Net for M=1, trained on TAIL labels to simulate situations where labels are not available. We used LayoutLMV3's feature extractor which uses Tesseract for training, and actual labels for testing the true performances to simulate real life. Note that while newer, more promising OCR solutions such as DocTr \cite{doctr2021} are available, these do not always support non-Roman alphabet languages or handwritten text, crucial for industrial application.

 \textbf{Assessment metric} We used \textbf{ANLS scores} (average normalized Levenshtein similarity score) proposed by \cite{ANLS} between the model being evaluated and the noisy labels (exact labels for CORD) as the metric of assessment  on the validation dataset, under the M=1 setup. Under the M=24 setup, we use \textbf{tree-edit distance} \cite{revisitingtreeeditdistance} on the JSON output.  We also compare the speeds of inference and costs. The cost of approaches are measured annually, assuming 100X documents(we withhold the exact numbers for confidentiality), Sonnet API call prices are taken as of April 2024, and the cost of TAIL labelling for 10K samples for LLaVA-Net and layoutLMV3 are included. The speed of inference is measured as documents processed per second for M=1. The results are shown in ~\cref{tab:rq1}. 
\begin{table*}[h]
	\small
		\label{sample-table}
		\centering
		\begin{tabular}{lllllllll}
			\toprule
			\textbf{Model} & \textbf{Merchant name} & \textbf{Amount} & \textbf{Date}  & \textbf{CORD, M=1}  & \textbf{CORD, M=24}& \textbf{Inference speed} & \textbf{Annual cost} \\
			\midrule
Sonnet Zero-Shot (TAIL)   & 50\%          & 70\%   & 76\% & 94\% & 93\% & 0.3 & \$1.25X & \\
LLaVA zero shot TAIL prompt    & 24\%          & 28\%   & 58\% & 48\%& 42\% & 1.5 & \$0.13X &   \\
LLaVA-noisy labels     & 46\%          & 63\%   & 68\%  & - & - & 1.5 & \$0.13X &   \\
LLaVA-exact labels     & -          & -   & -  & 82\% & - & 1.5 & 0.13X &   \\
LayoutLMV3-Tail & 41\%          & 58\% & -  &  42\%& - & 0.8 &0.2X &\\
LLaVA-Net & \textbf{52\%}          & \textbf{70\%}   & \textbf{83}\% &  84\%& 69\%  & \textbf{1.5} & \$0.2X &         
		\end{tabular}
		\caption{Performance Comparison: Sonnet, LLaVA, LLaVA-Net, LayoutLMV3}
    \label{tab:rq1}
	\end{table*}

 Sonnet zero-shot outputs (TAIL labels) showed ANLS scores of 50\%, 70\% and 76\% with the noisy labels. LLaVA zero-shot capability was poor compared to Sonnet. LLaVA trained on noisy labels performs significantly better that LLaVA zero-shot, learning the required output format. However, as the noisy labels do not always match with the text in the images, the vision component of LLaVA doesn't find the label in the image in many cases and this hampers fine-tuning. It performs worse than LLaVA-Net, showing the \textbf{lift due to TAIL labels.}. LLaVA-Net is able to show near-par performances with Sonnet on the internal dataset, and performs slightly better on the merchant name and amount questions, while being $\sim$ 85\% cheaper and 5X faster than Sonnet.  A surprising result was seen in CORD, where training on exact labels showed slightly worse performance that on TAIL labels, and in general LLaVA-Net performed poorer than Sonnet. We investigate this further in research question 3.
 
 While LayoutLMV3 trained on TAIL labels is able to learn to detect the positions of merchant names and invoice amounts, it is not able to perform as well as LLaVA-Net, and in research question 2, we investigate this further.

 \subsubsection{RQ 2: Why does LLaVA-Net outperform  layout-aware methods?}
Layout-aware methods are pre-trained and fine-tuned using masked language modeling and masked image modeling rather than language-based reasoning. They learn best when the fine-tuning dataset consists of samples with repeated formats. However, to take an example, the detection of the text "Giordano's Pizza" as merchant name, even if it is placed in an odd part of the invoice whose format is rare, requires reasoning. Since ExpenseQA consists of a wide variety of vendors and invoice templates, LayoutLMV3 is not able to extract information from rare document templates. 

To check this hypothesis, we looked at the performance of LayoutLMV3 on receipts from Uber and Lyft, two common vendors found in the dataset. LayoutLMV3 performed on par with Sonnet and LLaVA-Net on these vendors, as shown in \cref{tab:rq32}.
\begin{table*}[h]
	\small
		\label{sample-table}
		\centering
		\begin{tabular}{llll}
			\toprule
			\textbf{Model} & \textbf{Merchant name - Uber} & \textbf{Merchant name - Lyft} \\
			\midrule
Sonnet Zero-Shot (TAIL)   & 88\%          & 100\%    \\
LLaVA-Net & 88\%          & 100\%       \\
LayoutLMV3-Tail & 88\%          & 95\%          
		\end{tabular}
  
		\caption{Performance Comparison on common vendors: Sonnet, LLaVA-Net, LayoutLMV3-Tail}
    \label{tab:rq32}
	\end{table*}
This indicates that layout-aware models are able to effectively extract information from documents from common templates but fail in heterogeneous corpuses with rare templates. \textbf{The multimodal nature} and \textbf{reasoning capability} of LLaVA-Net allows it to use both templates and logic to extract information from expense documents.

LayoutLMV3 is also impacted by the quality of images, and of OCR. Since we used LayoutLMV3 assuming scenarios where labels are not available, in CORD dataset we used its tesseract-based feature extractor. This failed on poor-quality images leading to an ANLS of 42\%. However, on images where OCR successfully detected text in this window, the ANLS jumped to 72\%, showing the high dependence layout-aware methods have on underlying OCR used, as well as the superiority of Sonnet in vision-based tasks. 
Note that LayoutLMV3 performs well when provided OCR output and exact labels are used, showing micro-F1 of 96\% across all attributes in CORD \cite{perot2024lmdxlanguagemodelbaseddocument}.
 
 \subsubsection{RQ 3: Image quality impact on performance.}

\begin{table*}[h]
	\small
		\centering
		\begin{tabular}{llllll}
			\toprule
			\textbf{Model} & \textbf{$< 200$} & \textbf{$200-500$} & \textbf{$500-750$} & \textbf{$750-1000$} & \textbf{$\geq 1000$} \\
			\midrule
LLaVA-Net & 70\%  & 91\% & 90\% & 92\% & 92\%    \\
LayoutLMV3-Tail & 35\% & 45\%& 50\% & 60\%        & 33\%    \\
LayoutLMV3-Text in window & 67\% & 70\% & 85\% & 100\%       & 67\%    
		\end{tabular}
  
		\caption{Performance Comparison on differing image quality: LLaVA-Net, LayoutLMV3-Tail}
    \label{tab:quality}
	\end{table*}
To assess image quality, we used the variance of the Laplacian operator applied to an image. Intuitively, high-variance means sharp contrast between regions of high and low intensity, indicating low blur. For LayoutLMV3, we considered the general scenario, as well as receipts where its feature extractor successfully extracted text in its window.

As can be seen in \cref{tab:quality}, as focus improves, LLaVA-Net starts showing better results faster than LayoutLMV3, and at ranges with variances $\geq 200$, the ANLS exceeds 90\%, which is almost at par with Sonnet. For LayoutLMV3, in general the accuracy is poor. However, when the relevant text is present in its window, the accuracy goes up to 100\% for high-focus images.

This explains why LLaVA-Net performs so well on ExpenseQA, where $\sim$ 90\% images have Laplacian variance $\geq 200$, as opposed to CORD, where the number is $\sim 60\% $, where LayoutLMV3 fails due to heterogeneity.

 \subsubsection{RQ 4: Does LLaVA-Net replicate the errors made by Sonnet?}

\begin{figure}
  \centering
    \includegraphics[scale=0.15]{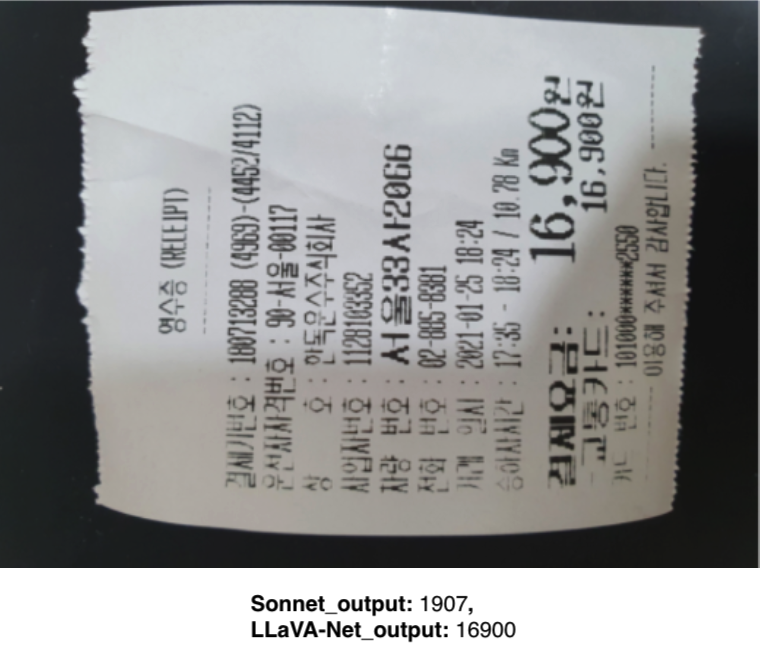}
\caption{LLaVA-Net correctly extracts amount from an invoice in portrait orientation, while Sonnet fails}
\label{portrait}
\end{figure}

The ANLS score between LLaVA-Net and Sonnet on the validation dataset in ExpenseQA for merchant name, amount and transaction date was 87\%,74\% and 82\%, respectively. In CORD, under M=1 setup the ANLS between LLaVA-Net and Sonnet outputs was 86\%, and tree-edit distance under M=24 was 87\%. Overall, ANLS and TED are less than 90 \%, indicating that LLaVA-Net isn't a simple replica of Sonnet. On further analyses of mismatches in names, we found that Sonnet had refused to answer in 3\% of samples, while LLaVA-Net had \textbf{zero refusals}. This is because LLaVA-Net was fine-tuned to generate text from the conditional distribution of the provided format. LLaVA-Net also correctly extracted merchant names in cases where Sonnet got confused by multiple logos/names in receipts (eg. Uber Eats receipts mentioning the name of restaurants fulfilling the order), or by the receipt being in landscape layout instead of the default portrait (\cref{portrait}). Sonnet also showed a 1.5\% rate of wrongly formatted output, generating text such as "Based on the information provided in the image, it seems like the receipt was generated by...". No such phenomenon was observed in LLaVA-Net, which generated outputs exclusively in the desired format.

\subsubsection{RQ 5: How can LLaVA-Net be used to detect potential fraud/abuse/wastage?}
\begin{figure}
  \centering
    \includegraphics[scale=0.25]{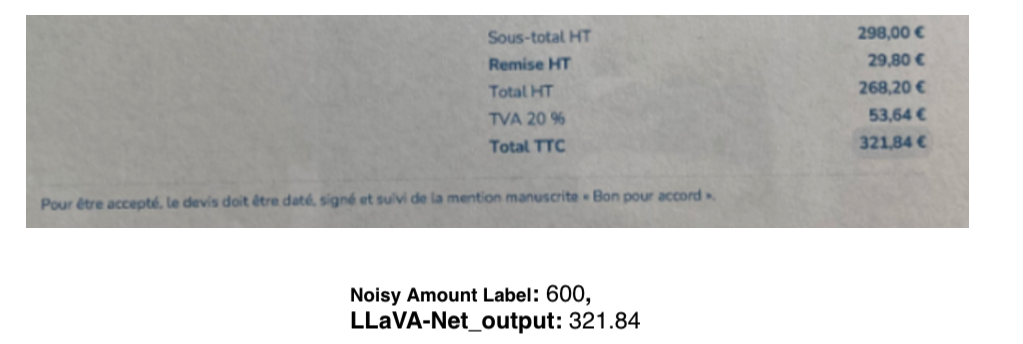}
\caption{Receipt where employee typed in 600 euros as the amount to be reimbursed, and LLaVA-Net detected the correct amount of 321.64 euros}
\label{overpaid_fin}
\end{figure}

In this section, we show a way in which potential fraud/abuse/wastage can be detected from LLaVA-Net output. 89 invoices were found in the expenseQA dataset where the extracted amount was lower than the amount typed in by employees. This indicates risk of overpayment, as the reimbursement depends on the amount typed in rather than the invoice submitted. \cref{overpaid_fin} shows an invoice with potential overpayment of $\sim$ 280 euros. Overall, the risk from these invoices was around \$2 per document, which becomes significant with scale.

Additionally, information such as extracted transaction date and location can be cross-checked with  available employee location data. Merchant name, product category can be used to check whether expenses align with spending policies. Aggregate spends on products/merchants, frequency of spends etc can be used for downstream anomaly detection.
\section{Conclusion}
In this paper, we have proposed TAIL, a mechanism to generate synthetic truth labels for VRDs without labels, and LLaVA-Net, a multimodel student model trained using response based knowledge distillation without using the teacher model's weights or training dataset. We applied LLaVA-Net to VRDU on internal and external datasets, and compared its performance with other methods that have shown promising results in literature.
\newline\textbf{Learnings} LLaVA-Net performs at par with SOTA Claude 3 Sonnet output under the condition that most of the images in the VRD corpus have decent image quality. If the images are poor, Sonnet's superior vision capabilities make it outperform LLaVA-Net. LLaVA-Net shows lower incidences of refusal and wrongly-formatted output, since finetuning leads it to generate text from the conditional distribution of corpus attributes. It outperforms layout-aware methods on heterogeneous document corpus due to its reasoning ability and superior vision capabilities, while layout-aware methods perform well under conditions of high homogeneity and high image quality. We have also shown that TAIL leads to more efficient downstream model training than noisy labels found in expense data. LLaVA-Net, comprising of 7B parameters, can be productionized using a single Sagemaker instance with one 20GB GPU (note that we used 8 GPUs solely for faster training), and is about $\sim$ 85\% cheaper than Sonnet, while being 5X faster. We also illustrated the usage of LLaVA-Net output to detect potential abuse/wastage/fraud through overpayment prevention.
\newline\textbf{Future Directions} As discussed in \cref{label_soln}, good practices such as ICL and COT were not used to improve latency and reduce cost, as given the huge volume of unlabelled receipts, both cost and latency were important concerns. In the future, prompting practices such as ICL, COT, automated prompt generation etc can be leveraged on top to further improve performance. We also discussed that if image quality is poor, LLaVA-Net shows suboptimal performance compared to Sonnet, which has superior vision. Super-resolution imaging can be used to improve resolution of images, and OCR output can be passed to the LMM inside prompts to tackle these cases, depending on budget and latency constraints.

{\small
\bibliographystyle{ieee_fullname}
\bibliography{egbib}
}\end{document}